%% file: preprint_main.tex
\documentclass[11pt]{journal}

\usepackage{amsmath,amssymb,amsfonts}
\usepackage{algorithmic}
\usepackage{graphicx}
\usepackage{textcomp}
\usepackage{xcolor}
\usepackage{float}
\usepackage{multirow}
\usepackage{siunitx}
\usepackage{footnote}
\usepackage{nameref}
\usepackage[margin=1in]{geometry}
\usepackage{authblk}
\usepackage[T1]{fontenc}

\usepackage{DenKr/patches/support_caption}%
\usepackage{caption}%
\usepackage{subcaption}
\usepackage{booktabs}

\usepackage{acronym}
\input{./chapter/acronyms.tex}


\usepackage[
	pdfborder={0 0 0},
	colorlinks=false,%
	urlcolor=black,%
	linkcolor=black,%
	citecolor=black,%
	filecolor=black,%
	breaklinks,%
	]{hyperref}%

\begin{document} 
\renewcommand{\tableautorefname}{Tab.} 
\renewcommand{\figureautorefname}{Fig.} 
\renewcommand{\sectionautorefname}{Section} 

\captionsetup{belowskip=-3pt}
\setlength{\textfloatsep}{10pt plus0pt minus6pt}

\title{Implementation Analysis of Collaborative Robot Digital Twins in Physics Engines
}

\author[2]{Christian König}
\author[2]{Jan Petershans}
\author[2]{Jan Herbst}
\author[2]{Matthias Rüb}
\author[2]{Dennis Krummacker}
\author[2]{Eric Mittag}
\author[1,2]{Hans D. Schotten}

\affil[1]{University of Kaiserslautern-Landau (RPTU), Germany\\
\texttt{\{lastname\}@rptu.de}}
\affil[2]{German Research Center for Artificial Intelligence (DFKI), Germany\\
\texttt{\{firstname\}.\{lastname\}@dfki.de}}
\date{}
\maketitle

%
%
%
%
\begin{abstract}
This paper presents a Digital Twin (DT) of a 6G communications system testbed that integrates two robotic manipulators with a high-precision optical infrared tracking system in Unreal Engine 5. Practical details of the setup and implementation insights provide valuable guidance for users aiming to replicate such systems, an endeavor that is crucial to advancing DT applications within the scientific community. Key topics discussed include video streaming, integration within the Robot Operating System 2 (ROS 2), and bidirectional communication. The insights provided are intended to support the development and deployment of DTs in robotics and automation research.\\
\end{abstract}
\textbf{This is a preprint of a work accepted but not yet published at the 7th International Congress on Human-Computer Interaction, Optimization and Robotic Applications (ICHORA 2025), technically co-sponsored by the IEEE Türkiye Section. The final version will be published in IEEE Xplore. Please cite as: C.König, J.Petershans, J.Herbst, M.Rüb, D.Krummacker, E.Mittag and H.D. Schotten: “Implementation Analysis of Collaborative Robot Digital Twins in Physics Engines,”
Proceedings of the 7th International Congress on Human-Computer Interaction, Optimization and Robotic Applications (ICHORA 2025), IEEE, 2025. }
\section*{Keywords}
Teleoperation, Game Engine, ROS 2, Unreal, LAM, Large Action Model, 6G

%
%
%
%

\section{Introduction} 
\input{./chapter/introduction.tex}

%
%
%
%

\section{Related Work} 
\input{./chapter/related_work.tex}

%
%
%
%

\section{The Real-World Setup}\label{sec:physical_setup}
\input{./chapter/real_world_setup.tex}

%
%
%
%

\section{The Digital Twin} 
\input{./chapter/dt_showcase.tex}

%
%
%
%
\section{Conclusion and Outlook}
\input{./chapter/conclusion.tex}

%
%
%
%
\section*{Acknowledgment}
\input{./chapter/acknowledgement.tex}

%
%
%
%
\bibliographystyle{IEEEtran}
\bibliography{references}
\end{document}

%% file: chapter/acronyms.tex
\newacro{5g}[5G]{Fifth Generation}
\newacro{6g}[6G]{Sixth Generation}
\newacro{as}[AS]{Authentication Server}
\newacro{atm}[ATM]{Automated Teller Machine}
\newacro{ai}[AI]{Artificial Intelligence}
\newacroplural{ais}[AIs]{Artificial Intelligences}
\newacro{aas}[AAS]{Asset Administration Shell}
\newacro{b5g}[B5G]{Beyond 5G}
\newacro{ban}[BAN]{Body Area Network}
\newacro{bsi}[\textit{BSI}]{\textit{Federal Office for Information Security}}
\newacro{bdr}[BDR]{Bit Disagreement Rate}
\newacro{bs}[BS]{Base Station}
\newacro{ca}[CA]{Certification Authority}
\newacro{cav}[CAV]{Connected Autonomous Vehicles}
\newacro{cc}[CC]{Common Criteria}
\newacro{cir}[CIR]{Channel Impulse Response}
\newacro{cr}[CR]{Challenge-Response}
\newacro{cpu}[CPU]{Central Processing Unit}
\newacro{cpps}[CPPS]{Cyber-Physical Production System}
\newacro{crl}[CRL]{Certificate Revocation List}
\newacro{csi}[CSI]{Channel State Information}
\newacro{crke}[CRKE]{Channel-Reciprocity Based Key Extraction}
\newacro{ctf}[CTF]{Channel Transfer Function}
\newacro{cotf}[COTF]{Commercial-off-the-Shelf}
\newacro{cmos}[CMOS]{Complementary Metal-Oxide-Semiconductors}
\newacro{cobot}[Cobot]{Collaborative Robot}
\newacro{dos}[DoS]{Denial-of-Service}
\newacro{ddos}[DDoS]{Distributed-Denial-of-Service}
\newacro{dds}[DDS]{Data Distribution Service}
\newacro{dna}[DNA]{Deoxyribonucleic Acid}
\newacro{dtls}[DTLS]{Datagram Transport Layer Security}
\newacro{dct}[DCT]{Discrete Cosine Transformation}
\newacro{dlt}[DLT]{Distributed Ledger Technology}
\newacro{dt}[DT]{Digital Twin}
\newacro{ds}[DS]{Digital Shadow}
\newacro{eal}[EAL]{Evaluation Assurance Level}
\newacro{ecc}[ECC]{Elliptic Curve Cryptography}
\newacro{ecg}[ECG]{Electrocardiogram}
\newacro{eeg}[EEG]{Electroencephalogram}
\newacro{embb}[eMBB]{enhanced Mobile broad-Band}
\newacro{emg}[EMG]{Electromyogram}
\newacro{eog}[EOG]{Electrooculography}
\newacro{enb}[eNodeB]{Evolved Node B}
\newacro{er}[ER]{Extended Reality}
\newacro{etsi}[ETSI]{European Telecommunications Standards Institute}
\newacro{fpga}[FPGA]{Field Programmable Gate Array}
\newacro{fdd}[FDD]{Frequency Division Duplexing}
\newacro{gdpr}[GDPR]{General Data Protection Regulation}
\newacro{gd}[G\&D]{Giesecke \& Devrient}
\newacro{h2m}[H2M]{Human-to-Machine}
\newacro{h2s}[H2S]{Human-to-Service}
\newacro{hmac}[HMAC]{Keyed-Hash Message Authentication Code}
\newacro{htc}[HTC]{Hologaphic-Type Communication}
\newacro{hotp}[HOTP]{HMAC-based One-time Password Algorithm}
\newacro{hsm}[HSM]{Hardware Security Module}
\newacro{ics}[ICS]{Industrial Control System}
\newacro{iacs}[IACS]{Industrial Automation and Control System}
\newacro{ioe}[IoE]{Internet of Everything}
\newacro{iiot}[IIoT]{Industrial Internet of Things}
\newacro{iot}[IoT]{Internet of Things}
\newacro{io}[I/O]{Input/Output}
\newacro{ic}[IC]{Integrated Circuit}
\newacro{id}[ID]{Identificator}
\newacro{ids}[IDS]{Intursion Detection System}
\newacro{irs}[IRS]{Intelligent Reflecting Surface}
\newacro{istn}[ISTN]{Integrated Space and Terrestrial Network}
\newacro{it}[IT]{Information Technology}
\newacro{itu}[ITU]{International Telecommunication Union}
\newacro{jcop}[JCOP]{Java Card Open Platform}
\newacro{kba}[KBA]{Knowledge Based Authentication}
\newacro{kdf}[KDF]{Key Derivation Function}
\newacro{lam}[LAM]{Large Action Model}
\newacro{led}[LED]{Light Emitting  Diode}
\newacro{lte}[LTE]{Long Term Evolution}
\newacro{ltea}[LTE-A]{Long Term Evolution Advanced}
\newacro{lr}[LR]{Linear Regression}
\newacro{los}[LoS]{Line of Sight}
\newacro{lorawan}[LoRaWAN]{Long Range Wide Area Network}
\newacro{mbb}[MBB]{Mobile Broadband}
\newacro{mfa}[MFA]{Multi-Factor Authentication}
\newacro{mcc}[MCC]{Mobile Cloud Computing}
\newacroplural{mcus}[MCUs]{Microcontroler Units}
\newacro{m2m}[M2M]{Machine-to-Machine}
\newacro{m2s}[M2S]{Machine-to-Service}
\newacro{mimo}[MIMO]{Multiple Input Multiple Output}
\newacro{mmimo}[mMIMO]{massive Multiple Input Multiple Output}
\newacro{ml}[ML]{Machine Learning}
\newacro{mulc}[mULC]{massive Ultra-Reliable Low-Latency Communication}
\newacro{mmtc}[MMTC]{massive Machine Type Communication}
\newacro{mmg}[MMG]{Mechanomyogram}
\newacro{multos}[MULTOS]{Multii-Application Smart Card Operating System}
\newacro{mux}[MUX]{Multiplexer}
\newacro{mnc}[MNC]{Mobile Network Code}
\newacro{me}[ME]{Mobile Environment}
\newacro{mac}[MACs]{Message Authentication Codes}
\newacro{mocap}[MoCap]{Motion Capture}
\newacro{mbse}[MBSE]{Model Based System Engineering}
\newacro{ngmn}[NGMN]{Next Generation Mobile Network}
\newacro{nic}[NIC]{Network Interface Controller}
\newacro{nist}[NIST]{National Institute of Standards and Technology}
\newacro{oath}[OATH]{Open Authentication}
\newacro{ocra}[OCRA]{\ac{oath} Challenge-Response Algorithm}
\newacro{ocsp}[OCSP]{Online Certificate Status Protocol}
\newacro{otp}[OTP]{One-Time Password}
\newacro{pap}[PAP]{Password-Authentication-Protocol}
\newacro{physec}[PhySec]{Physical Layer Security}
\newacro{pfs}[PFS]{Perfect Forward Secrecy}
\newacro{pin}[PIN]{Personal Identification Number}
\newacro{pkc}[PKC]{Public Key Cryptography}
\newacro{pki}[PKI]{Public Key Infrastructure}
\newacro{ppg}[PPG]{Photoplethysmography}
\newacro{prng}[PRNG]{Pseudo Random Number Generator}
\newacro{puf}[PUF]{Physically Unclonable Function}
\newacroplural{pufs}[PUFs]{Physically Unclonable Functions}
\newacro{pla}[PLA]{Physical Layer Authentication}
\newacro{plm}[PLM]{Product Lifecycle Management}
\newacro{qr}[QR]{Quick Response}
\newacro{rat}[RAT]{Radio Access Technology}
\newacro{radius}[RADIUS]{Remote Authentication Dial-In User Service}
\newacro{ram}[RAM]{Random-Access Memory}
\newacro{ran}[RAN]{Radio Access Networks}
\newacro{rf}[RF]{Radio-Frequency}
\newacro{rfid}[RFID]{Radio-Frequency Identification}
\newacro{ris}[RIS]{Reconfigurable Intelligent Surface}
\newacro{rng}[RNG]{Random Number Generator}
\newacro{ro}[RO]{Ring-Oscillator}
\newacro{rom}[ROM]{Read-Only Memory}
\newacro{rs}[RS]{Reed-Solomon}
\newacro{rsa}[RSA]{Rivest-Shamir-Adleman}
\newacro{rssi}[RSSI]{Received Signal Strength Indicator}
\newacro{rsrp}[RSRP]{Reference Signal Received Power}
\newacro{re}[RE]{Resource Elements}
\newacro{ros2}[ROS~2]{Robot Operating System~2}

\newacro{sdn}[SDN]{Software-Defined Network}
\newacro{sdr}[SDR]{Software-Defined Radio}
\newacro{seccos}[SECCOS]{Secure Chip Card Operating System}
\newacro{sip}[SIP]{Session Initiation Protocol}
\newacro{skg}[SKG]{Secret Key Generation}
\newacro{sram}[SRAM]{Static Random Access Memory}
\newacro{srs}[SRS]{Software Radio Systems}
\newacro{starcos}[STARCOS]{Smart Card Chip Operating System}
\newacro{sha}[SHA]{Secure Hash Algorithm}
\newacro{se}[SE]{Static Environment}
\newacro{svm}[SVM]{Support Vector Machine}
\newacro{sysml}[SysML]{System Modeling Language}
\newacro{tcg}[TCG]{Trusted Computing Group}
\newacro{tpm}[TPM]{Trusted Platform Module}
\newacro{tls}[TLS]{Transport Layer Security}
\newacro{trng}[TRNG]{True Random Number Generator}
\newacro{tsn}[TSN]{Time-Sensitve Networking}
\newacro{tofu}[TOFU]{Trust On First Use}
\newacro{tufu}[TUFU]{Trust Upon First Use}
\newacro{totp}[TOTP]{Time-based One-time Password Algorithm}
\newacro{uav}[UAV]{Unmanned Arial Vehicles}
\newacro{usb}[USB]{Universal Serial Bus}
\newacro{usrp}[USRP]{Universal Software Radio Peripheral}
\newacro{uhd}[UHD]{USRP Hardware Driver}
\newacro{usim}[USIM]{Universal Subscriber Identity Module}
\newacro{ue}[UE]{User Equipment}
\newacro{urllc}[URLLC]{Ultra-Reliable Low-Latency Communication}
\newacro{ulbc}[ULBC]{Ultra-Reliable Low-Latency Broadband Communication}
\newacro{umbb}[uMBB]{ubiquious Mobile Broadband}
\newacro{ummimo}[UM-MIMO]{Ultra-Massive MIMO}
\newacro{urdf}[URDF]{Unified Robot Description Format}
\newacro{uml}[UML]{Unified Modeling Language}
\newacro{vlc}[VLC]{Visible Light Communication}
\newacro{VRel}[VRel]{Virtual Relative}
\newacro{warp}[WARP]{Wireless open-Access Research Platform}
\newacro{wysiwyg}[WYSIWYG]{What You See Is What You Get}

\newacro{xr}[XR]{Extended Reality}

%% file: chapter/introduction.tex
\label{sec:introduction}

Digital Twinning is an emerging concept that finds its application in many different domains like manufacturing, energy, or automotive industries~\cite{OmniverseShowCase, UnityShowCase, UnrealShowCase}.
With the advent of \acp{lam}, \acp{dt} are becoming increasingly relevant. \acp{dt} facilitate the training of \acp{lam}—advanced \ac{ai} models designed for complex decision-making and control—while minimizing risks to real-world systems.
\acp{dt} provide a virtual representation of physical systems, enabling simulation, analysis, and optimization without the risks associated with real-world experimentation. Specifically, \acp{dt} allows the testing of \ac{lam} behavior in rare and unforeseen scenarios.

Mashaly presents the manufacturing sector as use cases for \ac{dt}, where they can be used to test, control, and optimize the production plant without disrupting operations.
They further highlight the healthcare sector, training, and simulation environment for surgeons, where an ultra-reliable low latency network connection is of indispensable need~\cite{MASHALY2021299}.
Halúsková highlights the benefits of \acp{dt} for a smart city, which can be used to solve infrastructure and transport network problems~\cite{HALUSKOVA20231471}.
From the fundamentally different challenges of the individual areas, it is evident that each \ac{dt}, must be carefully designed to align with its objectives and the resources required to accomplish them.

\begin{figure}[H]
    \centering
    \includegraphics[width=0.7\linewidth]{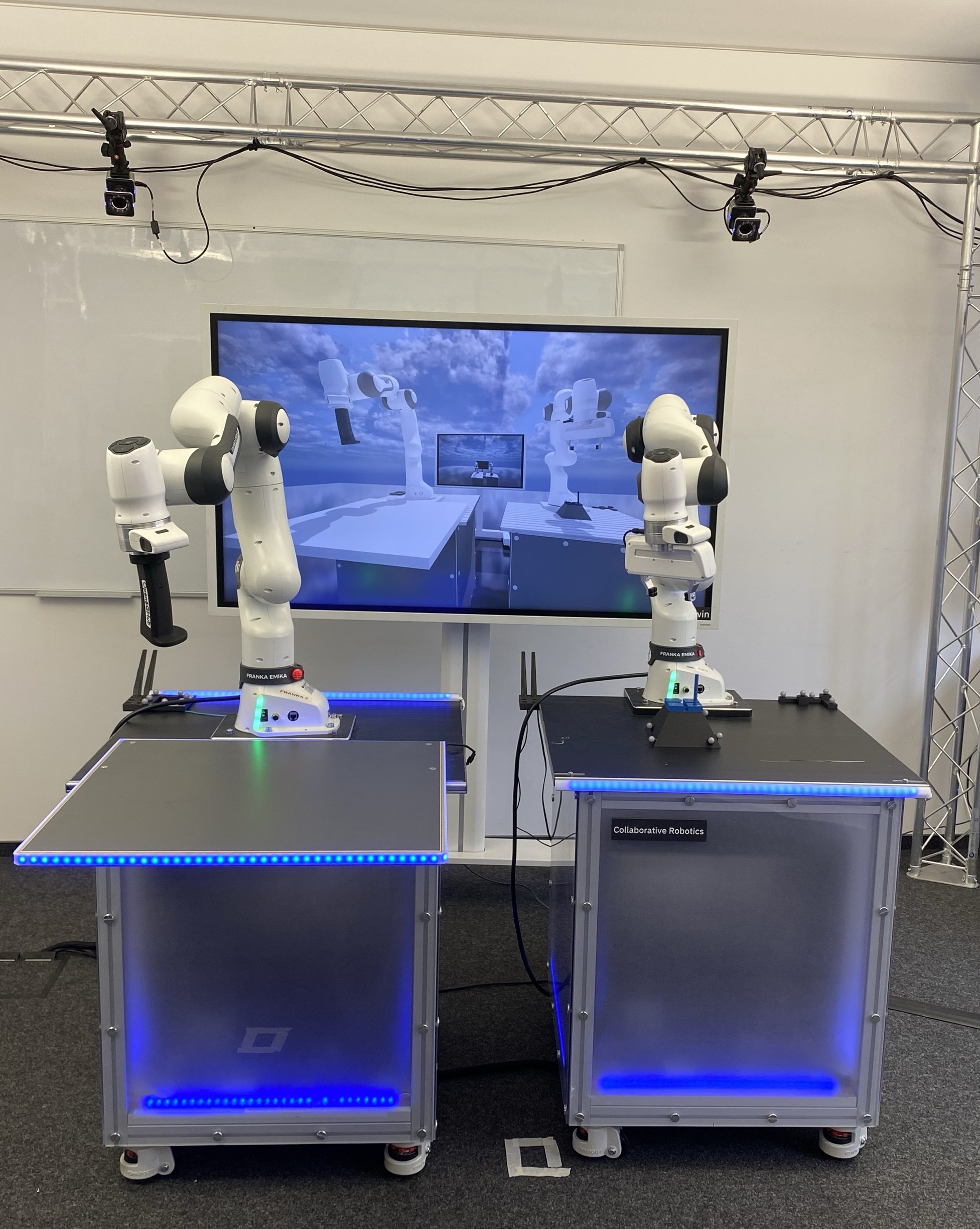}
    \caption{Image of the teleoperation test environment for communication systems, with its \ac{dt} visualized in the background. The cobot on the left represents the leading system (Leader), while the cobot on the right represents the following system (Follower).}
    \label{fig:testbed_photo}
\end{figure}

\acp{lam} should be able to react appropriately to unforeseen and rare circumstances. Through the implementation inside a \ac{dt}, not only safe training and testing of these models under realistic conditions is possible, but also regarding real-world cases, the system obtains a holistic view of the controlling setup. Implementing \acp{lam} within \acp{dt} will present lower risks compared to real-world implementations—eliminating the possibility of equipment damage or personnel injury during testing—and can achieve cost efficiency through reduced need for physical prototypes. Additionally, it enhances data generation for refining \ac{ai} models and improves safety and compliance by allowing thorough testing of safety protocols and adherence to regulations before real-world deployment. Beyond testing and training, the main value of the combination of \ac{lam} and \ac{dt} is during actual deployment as a monitoring medium. The combination offers a holistic view of the entire system. While simplifying the implementation of interfaces for \acp{lam} and still providing an encapsulated environment, it enhances real-time monitoring and predictive maintenance capabilities, thereby improving operational efficiency and decision-making processes. 

With arising advancements in physics engines and \ac{ai} frameworks, increasingly detailed and accurate simulations are enabled, offering significant potential for virtual training of real-world scenarios. Advanced platforms, such as NVIDIA`s Omniverse and Cosmos\footnote{2025-01-16, \url{https://www.nvidia.com/de-de/ai/cosmos/}}, demonstrate early implementations of these capabilities, providing the foundation for the creation of virtually generated environments through the application of \ac{ai}-driven simulation. Although such virtual platforms offer significant advantages—including risk-free experimentation, scalable data aggregation, and the potential for adaptive learning—effective deployment in collaborative robotics and teleoperation also necessitates a robust interface with the physical domain. Clear protocols for data exchange and control are essential to ensure a seamless transition to real-world applications.

Thus, this work focuses on the implementation of a \ac{dt} for collaborative robotics. Therefore, a flexible robotic teleoperation test environment with two collaborative robots (cobots) from Franka Robotics (Franka Research 3) is used, previously presented in~\cite{Petershans2024_RoboticTeleoperationRealWorld}. The laboratory setup, which incorporates an infrared sub-millimeter precision position tracking system, is presented in \autoref{fig:testbed_photo}. This setup provides a foundational framework for the implementation and development of an accurate \ac{dt} of the laboratory environment.
The technical integration of communication between the components for this exemplary application provides a critical basis for development and improvement of similar systems. The cobots are modeled in Unreal Engine 5, leveraging their detailed specification and model structures to ensure accurate simulation. Furthermore, practical insights and guidance are provided on key topics including video streaming, integration with the middleware \ac{ros2}, and bidirectional communication.

The remainder of this document is structured as follows: \autoref{sec:related_work} gives a general overview of related work, introduces relevant terminology, and positions this work within established concepts. In \autoref{sec:physical_setup} the physical system, covering essential components such as the cobots, the position tracking system, and communication infrastructure are introduced. \autoref{sec:dt_showcase} provides in-depth insights of the development of the \ac{dt}, discussing key aspects including engine selection, the construction of the virtual cobots, network integration, and tracking solutions. Finally, the findings are summarized in \autoref{sec:conclusion} and potential directions for future research are outlined.

%% file: chapter/related_work.tex
\label{sec:related_work}
While existing literature explores \acp{dt} for robotic systems in Unreal Engine, it primarily focuses on specific use cases but lacks general guidance on the overall implementation process. This work tries to close this gap by describing in detail the limitations of the components and proposing alternative solutions.
\textit{Coronado~et al.} list commonly used frameworks and software components for the implementation of \acp{dt}~\cite{Coronado2023}. However, the authors mainly list the utilized tools based on their popularity in recent publications, without elaborating incompatibility problems during implementation.

According to the technical report 22.804 by 3GPP's industrial collaborators, collaborative robotics, eHealth, and telecare are identified as some of the most important use cases to be supported by communication systems.
The \ac{etsi} adopted the performance characteristics of these use cases into technical specifications for service requirements, such as \ac{etsi}~TS~122~104, and \ac{etsi}~TS~122~261, which include applications like telesurgery and collaborative robotics.
This reflects the telecommunication industry's strong confidence that these functionalities will be widely adopted in the application market.

Numerous works have been published on the application of \ac{dt} technologies in manufacturing. One key takeaway is its utility in controlling robots, e.g. for collaborative tasks or synchronization~\cite{10188847}.
The \ac{itu} also categorizes \acp{dt} as an important development in the context of distributed systems. 
Several ongoing studies and published recommendations addressing network-related aspects are managed by their standardization sector under the Y series (e.g., Y.DTN, Y.3090)~\cite{ITU-T}.

The healthcare sector is also actively working on implementing teleoperation with robotic systems, e.g. for surgery, care, or manufacturing. As responsiveness and reliability are indispensable in this domain~\cite{Petershans2024_RoboticTeleoperationRealWorld}, digital twinning is considered a key aspect for enhancing such systems.
Despite the advancements in the concept of \acp{dt}, a critical gap remains: the lack of a universally accepted framework for development and a standardized platform for commissioning. This technology is still in its phase of comprehensive definition and requirement specification. Consequently, existing solutions are predominantly proprietary~\cite{10633478,10717667,10241448}.
Hence, this work presents one step towards unification between the advancing concept of digital twinning and relevant application domains, such as robotics and healthcare.

\subsection{Terminology}

\subsubsection{Level of Integration}
\textit{Krummacker~et al.} proposed a broad term with a continuous spectrum for all virtual representations of some physical counterpart, called the \ac{VRel}. A \ac{VRel} may have different levels of data flow and connectivity to its physical counterpart~\cite{krummacker_digital_2023}.
\textit{Kritzinger~et al.} proposed a taxonomy for these levels. Hereby the first level is called the Digital Model. The Digital Model is only a digital representation of the physical counterpart without any form of connectivity between these two worlds. The second is called the \ac{ds}. That is a uni-directional virtual relative, that reflects the state of the physical counterpart. The highest level is called \ac{dt} which consists of a bidirectional communication~\cite{kritzinger_digital_2018-3}. This hierarchy is increasingly adopted in research~\cite{singh_digital_2021,cruz_systematic_2022}. 

\textit{Negri and Abdel-Aty} presented a study aiming to address existing inconsistencies in terminology associated with Metaverse, \ac{dt}, and Digital Thread. They conceptualize the Digital Thread as the seamless integration and exchange of information across the product lifecycle~\cite{negri_clarifying_2023}.
\textit{Zhang~et al.}, on the other hand, describe the Digital Thread as a structured set comprising the native model—encompassing 3D Models, behavioral representations in \ac{sysml}, and other related elements—the surrogate model, which serves as an abstraction of the native model using \ac{uml}, and the relationship between these models~\cite{zhang_literature_2024}.

Beyond merely reflecting the virtual counterpart, the proposed laboratory environment enhances the \ac{dt} by incorporating contextual information about the cobots' surroundings and positions. This supplementary data enables the \ac{dt} environment to automatically generate and transmit commands from the \ac{dt} to the respective physical counterpart. Consequently, due to this bidirectional communication and enrichment, the system presented in this work can be classified as \ac{dt}.

\subsubsection{Digital Twin Modeling Concepts}
Different approaches exist for designing and implementing a \ac{dt}. 
Physics-based models rely on established physical equations, incorporating structural information about the physical system under investigation. This approach results in models that exhibit high interpretability and explanatory power, requiring only small amounts of data. Yet, implementation for complex systems requires substantial expertise. Additionally such models may be constrained in accuracy due to incomplete understanding or simplification of certain intricate processes and challenges posed by computational complexity. 
On the other hand, data-driven models can effectively represent complex systems without relying on simplified equations or requiring comprehensive knowledge of all underlying processes within the system. However, to achieve an adequate representation, a comprehensive dataset is essential---one that is sufficiently large, adheres to high quality requirements, and encompasses all potential states of the system. Furthermore, the inherent nature of these models, operating without constrains imposed by underlying physical principles, can yield in results potentially contradicting fundamental physical laws
~\cite{Kapusuzoglu2020_PhysicsInformedHybridMachine}.

It is therefore crucial to identify the primary purpose of the \ac{dt}. For instance, if the \ac{dt} is intended primarily as controlling environment enriched with additional information, a physics-based model may be adequate. 
In a \ac{dt} environment designed for the learning or training of a physical system, it is essential to not only ensure an accurate reflection of the system’s state but also to facilitate the seamless application of information generated within the \ac{dt} to the physical world. This capability is particularly important for identifying unforeseen events, highlighting the benefits of a data-driven approach. The adoption of a hybrid model, as proposed by \textit{Daw~et al.}, offers a further opportunity by harnessing the strengths of both methodologies effectively~\cite{Daw2022_PhysicsGuidedNeuralNetworks}.

The utilization of a physics-based model for the \ac{dt} of the presented teleoperation scenario establishes a robust foundation for accurate system simulation and analysis. However, to further advance the system’s capabilities and adaptability, the integration of \ac{ai}-powered strategies into the \ac{dt} presents significant potential for improvements.

\subsection{Asset Administration Shell—AAS}
The \ac{aas} represents a \ac{plm} framework that standardizes interoperability and organizes digital realizations of physical assets hierarchically, valuable for lifecycle management and industrial integration~\cite{__SpecificationAssetAdministration}.
Thus, the \ac{aas} can be interpreted as a structured framework contributing to the broad concept of the Digital Thread by standardizing the representation and integration of product-related data across its lifecycle.

This work focuses on a \ac{dt} test environment for cobotic research, emphasizing bidirectional communication, multi-modal model testing, and integration with tools like Unreal Engine (5.3) and \ac{ros2}. Although the discussed functionalities---such as 3D visualization, robot-environment interaction, and system control---are compatible with integration into an \ac{aas} framework, this work primarily focuses on their precise implementation and organization at the application level within \ac{ros2} and the 3D visualization environment.

Nevertheless, \ac{aas} holds potential for future enhancements of the proposed \ac{dt} framework, particularly in addressing advanced objectives like \acp{lam}, which demand robust frameworks capable of managing complex and unpredictable scenarios. Aligning the system with full lifecycle interoperability positions \ac{aas} as a complementary tool to enhance modularity and scalability, facilitating the transition towards Type3 \ac{aas} in the long-term perspective~\cite{mike_rauh.etal_2022_AIAssetManagement}.


%% file: chapter/real_world_setup.tex
\label{sec:real_world_setup}
\begin{figure*}[t!]
    \centering
    \includegraphics[width=0.9\linewidth]{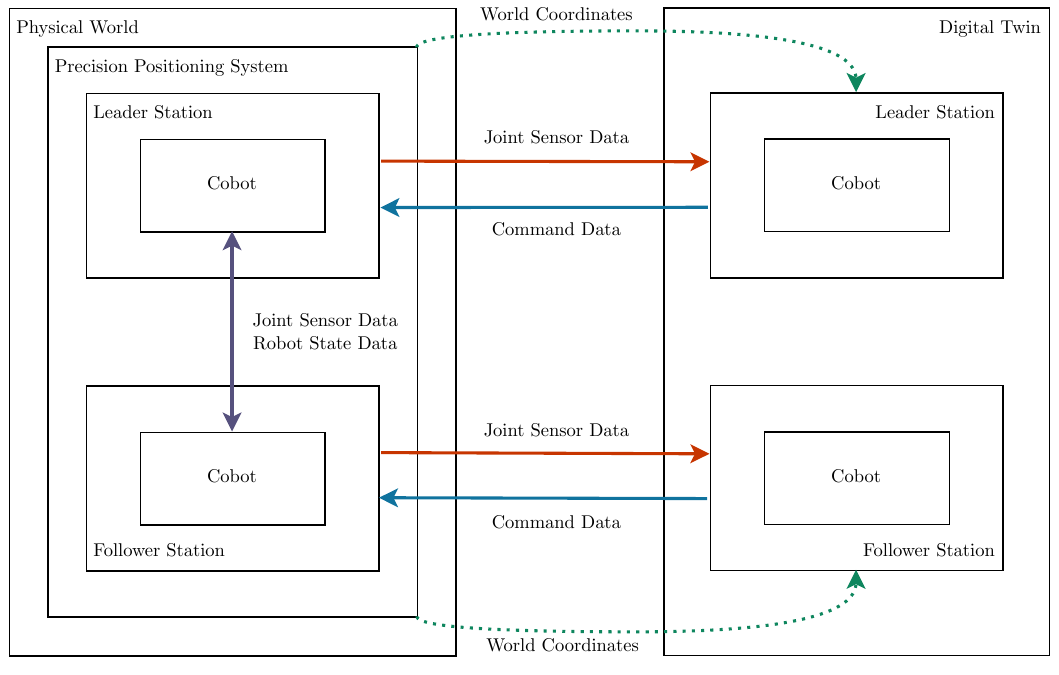}
    \caption{Abstraction of data exchange in the setup. The bidirectional communication based on \ac{ros2} of the teleoperation setup is indicated by the violet arrow. Communication between the cobots and their respective \ac{dt} is established via UDP and TCP connections. The joint sensor data is extracted from respective \ac{ros2} messages and transmitted via UDP connections (orange arrow), while command data from the \ac{dt} is transmitted via TCP connections (blue arrow) and converted into \ac{ros2} messages. Along with coordinates of the cobots provided by the precision positioning system (green dashed arrow), all data is merged in the \ac{dt} of the respective cobot.}
    \label{fig:data_exchange}
\end{figure*}

Latency, moreover its consistency, is a critical aspect in real-time applications involving robotics. Not only in processing and controlling a robot but also in the communication and transmission of relevant data. Therefore a test environment for communication technologies has been developed by \textit{Petershans~et al.}, using two Franka Research 3 cobots from Franka Robotics. The cobots are configured in a leader-follower setup for teleoperation demonstrations, where the follower mirrors the motion of the leader cobot~\cite{Petershans2024_RoboticTeleoperationRealWorld}. This setup allows augmentation by functional components, such as \ac{ai} and \acp{dt}, as well as other technologies such as \ac{mocap}. 

The following chapter explores the integration of the teleoperation setup with a precision positioning system. After describing these two parts of the overall setup, the communication between the components is discussed.

\subsection{The Teleoperation Setup}

The leader can be hand-guided by a person, while the follower, mirrors the leader's motion. Force applied externally to the follower results in a directly generated force feedback on the leader. The controller architecture is based on the open-source \textit{teleopJointPDExampleController} by Franka Robotics using joint-impedance-control and the ROS framework. Each cobot is mounted on a mobile machinery table, representing a local (leader) and a remote (follower) station, each operating on a dedicated computer system. Necessary data for the teleoperation setup like joint position and velocity or force applied to the end effectors is transmitted between the controllers. Communication between the cobots and their respective computer controlling it is performed with an update rate of \SI{1000}{Hz}. For evaluating the real-time capability of different communication systems, this frequency is also used for the teleoperation controller itself~\cite{Petershans2024_RoboticTeleoperationRealWorld}. For better guiding the leader, a joystick is mounted to it instead of an end effector, and a Franka Hand is attached to the follower, enabling it to grasp objects. This gripper can be controlled via a button on the joystick, communicating via Wi-Fi with the middleware.

The computers controlling the cobots run on Debian 12.8, using the PREEMPT\_RT kernel v6.1.0-27-rt. To facilitate augmentation of the teleoperation setup, the controllers have been adapted to the architecture of \ac{ros2} and run inside Docker containers on Ubuntu 22.04. \ac{ros2} is a flexible and efficient framework for creating robotic applications based on nodes and publishing and subscribing to topics and services. Any node can publish messages to a specific topic, and subscribers to this topic automatically receive the relevant information. This abstraction greatly simplifies communication within the framework, particularly in distributed systems. As its underlying middleware for data exchange \ac{ros2} relies on \ac{dds}, providing real-time capability and a robust messaging system. 

\subsection{Precision Positioning System}

Since the cobots have no information about their environment or location, an auxiliary solution is required, where an application needs spatial information, to map their positions. Especially in dynamic environments contextual awareness and a clear understanding of relative positions to other cobots or assets is essential. 

To provide precise coordinates of the two cobot stations, which can be positioned flexibly, an optical \ac{mocap} system from Vicon is used. Assuming this system is correctly calibrated, it offers motion tracking with a spatial resolution of up to \SI{0.1}{mm}. Eight infrared cameras are installed on a truss system with a base area of 3~$\times$~4\;m recording infrared light reflected by retro-reflective markers. Due to the optical nature of the tracking system, stable environmental conditions are essential for high quality measurements, especially in dynamic mixed reality environments, where reflective surfaces may interfere with the reflections of the actual markers. To prevent issues arising from a cluttered environment, maintaining steady lighting and either concealing or removing reflective surfaces is crucial.

Considering this foundation of stable environmental conditions, the teleoperation setup is located within the mapped volume of the \ac{mocap} system. 
To ensure a clear identification of the cobot stations, these are equipped with (for the \ac{mocap} system) unique 3d printed marker objects, allowing the stations to be tracked and distinguished by the software, Tracker 3.10. Thus, the positions of the cobots are known and can be forwarded to the \ac{dt} environment, 
enabling the \ac{dt} to dynamically adapt to changes in the real world.

\subsection{Communication}
In \autoref{fig:data_exchange} an abstract visualization of the communication between the various components of the system is depicted. On the left, the physical setup is showcased, consisting of the teleoperation setup and the precision positioning system, while on the right the \ac{dt} environment is displayed. The teleoperation setup, positioned within the mapped volume of the \ac{mocap} system, exchanging relevant data like joint position, velocity and measured external force between the leader and follower computer bi-directionally. This setup relies on \ac{ros2} for control and communication, reprensented by the violet arrow, offering a high level of abstraction for the user. 

However, Unreal Engine 5 does not include a native \ac{ros2} plugin, necessitating an alternative, cross-platform approach. Since rosbridge is not suitable for real-time applications with high data and transmission rates~\cite{Coronado2024_PathIndustry50}, a custom implementation has been developed. This implementation retrieves data from relevant topics, such as joint sensor information, at the \ac{ros2} level and establishes a UDP connection between the computers controlling the cobots and the \ac{dt} machine (orange arrows). The display presenting the \ac{dt} operates at an update rate of 60 Hz. Consequently, the UDP stream directed towards the \ac{dt} is throttled to 60 Hz, ensuring efficient data transmission and minimizing unnecessary processing overhead.
Conversely, command data generated by the \ac{dt} is transmitted via a TCP connection to the computers managing the teleoperation setup and then converted into \ac{ros2} messages, which is displayed by the blue arrows.

The coordinates of marker objects is transmitted from the \ac{mocap} computer to the \ac{dt} machine by a UDP stream, represented by the green dashed arrows. All the information gathered by the individual systems is merged by the \ac{dt} environment into the virtual representations of the two cobot stations.

%% file: chapter/dt_showcase.tex
\label{sec:dt_showcase}
This chapter presents a comprehensive exploration of developing a \ac{dt} using Unreal Engine as the core platform. The opening section explains the rationale behind selecting Unreal as the foundation, followed by a detailed description of how the robot twin is constructed within the Unreal environment. 
After establishing the \ac{dt}`s basic framework, networking and contextualization within the real world is discussed. Finally, the chapter concludes by exploring additional features and functionalities exceeding the core replication of a robot’s appearance and basic operations.

\begin{figure}[!ht]
    \centering
    \includegraphics[width=0.9\linewidth]{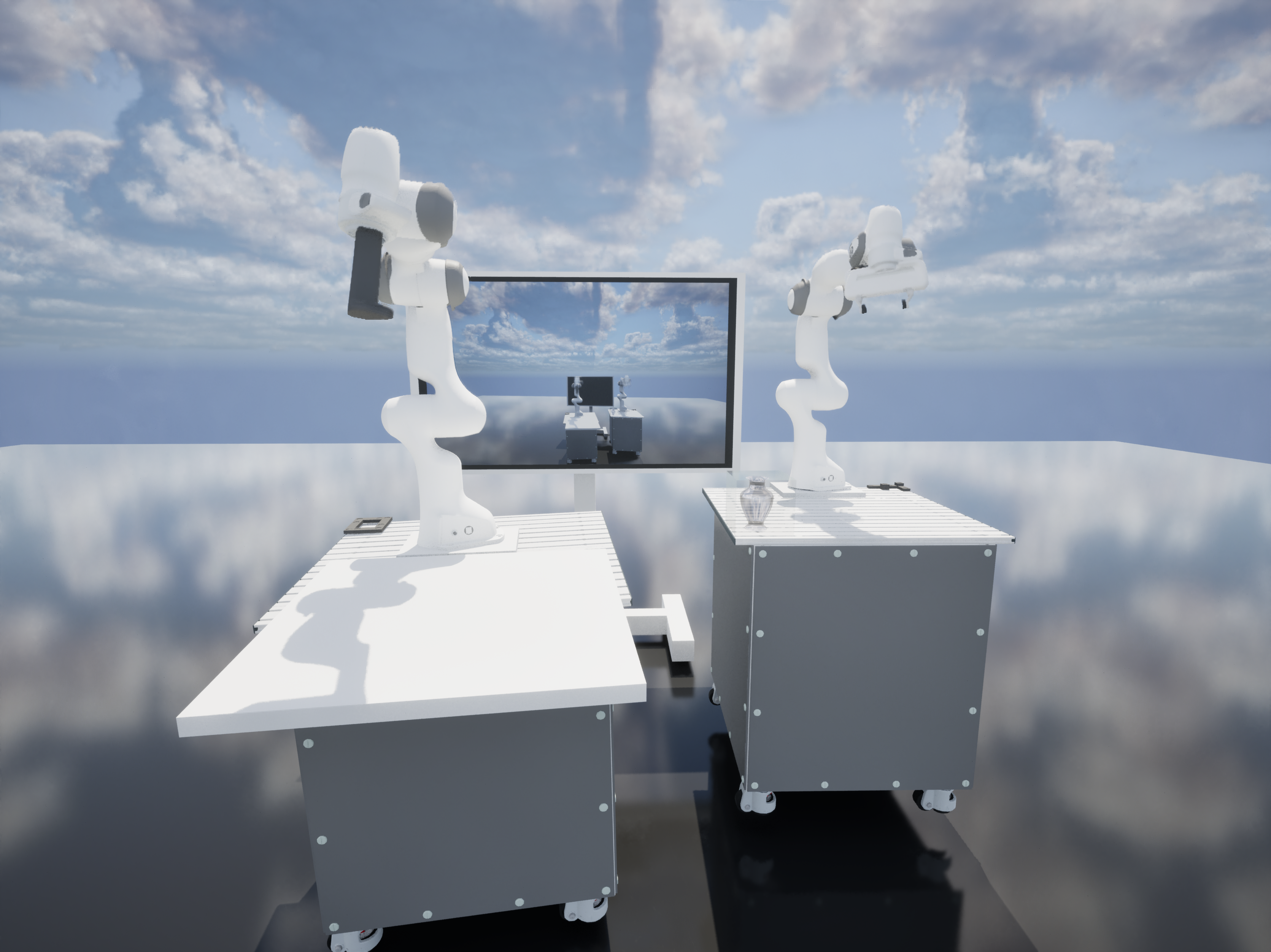}
    \caption{A screenshot of the \ac{dt} environment displays the system representation, with the leader station on the left and the follower station on the right. In the rear corners of the stations, the individual marker objects are depicted. At the front left of the follower station, a vase is visible, accompanied with a bounding box delineating a prohibited area. In the background, the visualized scene from a virtual camera is presented.}
    \label{fig:francaScreenshot}
\end{figure}

\subsection{Engine Selection}
In the domain of visualizable \acp{dt}, a range of simulation engines are available. One longstanding and frequently debated question is whether Unity is the most suitable choice for such applications. While Unity is well-established, and can achieve high-fidelity visuals through its various render pipelines, it often requires substantial effort and fine-tuning to produce truly remarkable results.

A notable newcomer to the market is NVIDIA Omniverse with Isaac Sim, offering an extensive amount of inter-connectivity features. Unlike Unity, Omniverse relies exclusively on ray tracing rather than rasterization. While this approach can produce highly realistic visuals, it comes at the cost of significantly higher computational requirements and may introduce lag—an undesirable outcome for teleoperation. Although Isaac Sim provides a wide range of tools for \ac{ai}, its relative novelty leads to stability limitations compared to more established simulation platforms. Stability is a critical factor in teleoperation applications, where reliable and consistent performance is paramount.

In contrast, Unreal Engine 5 offers sophisticated features, such as \textit{Lumen} for advanced lighting with global illumination and \textit{Nanite} for effortless level-of-detail management and mesh optimization, though it demands higher computing resources than e.g. Unity. It supports both rasterization and ray tracing, enabling fine-grained control of it's performance, and is able to render photorealistic environments while also handling large-scale \acp{dt}.

Thus, Unreal Engine (version 5.3) is selected for its balance between high visual fidelity and efficient performance. Furthermore, its stability and robustness make it a reliable choice for the intended application.

For this work, the \ac{dt} is running on a Windows computer with an AMD Ryzen 9 7950X CPU, \SI{128}{GB} of RAM, and an NVIDIA RTX 3090 graphics card.

\subsection{Constructing a Robot}
\subsubsection{Creating the Model}

For modeling a robot, detailed information about its links and joints is required, including their dimensions, positions, and other parameters. Using the \acp{urdf} file of the cobots, offering a comprehensive blueprint encompassing relative positions, meshes, and materials for every component, the physical configuration of the robot can be accurately replicated in a virtual environment. Since the primary objective is twinning the teleoperation rather than a full physics simulation, representing the cobots as a hierarchy of linked components reflecting each joint state (position, velocity and effort) is sufficient.

Unreal Engine does not include native support for \ac{urdf} files, necessitating the use of alternative methods. However, it supports file formats for 3D models such as \textit{.fbx}, \textit{.obj}, and \textit{.gltf}, as well as \textit{.usd*} (currently in beta). Consequently, an intermediate software tool is required to convert \textit{.urdf} files into a compatible format. Tools such as NVIDIA Omniverse or Blender can be employed for this purpose.

Although Unreal Engine offers a beta plugin for handling \textit{.usd*} files, this approach has proven inadequate for creating robot models. During testing, meshes exported from Omniverse in the \textit{.usd*} format exhibited significant deformation, while scenes exported from Blender in the same format resulted in a combined single mesh.
Similarly, \textit{.fbx} files exported from Blender versions $\leq$ 4.0 suffered from severe degradation. In contrast, Blender 3.6 or 3.3 produced more consistent results, though \textit{.fbx} files from both Blender and Omniverse Isaac Sim either combined all meshes into a single entity or projected them to a single anchor coordinate, necessitating manual adjustment.
Likewise, all meshes are combined into a single entity for \textit{.obj} files, making them unsuitable for complex mechanisms such as a multi-axis robot. 

Among the tested formats, only \textit{.gltf} files successfully preserved the robot's structure. This format provides textured meshes with correctly positioned links as individual meshes. 
To create a model in this file format, Phobos presents a suitable solution. This is an open-source plugin developed at the Robotics Innovation Center of German Research Center for Artificial Intelligence in Bremen, Germany, for the creation of robots within a \ac{wysiwyg} editing environment, supporting \ac{urdf} files. Using Blender's standard export capabilities, the robot model can be converted to desired format~\cite{phobos}.

To prepare a robot model for use in Unreal using Phobos, certain modifications are required. Collision shapes must be removed as they interfere with the visual fidelity of the \ac{dt}. Since movement calculations, including self-collision avoidance, are performed on the robot's onboard computer, collision shapes designed to prevent self-intersection are redundant.
Additionally, the skeleton layers of the links have to be adjusted in the Object Properties Panel to prevent visually distracting skeleton links from appearing in Unreal. These skeleton links are not functional for applying rotations or transformations to the joints. 

After applying these adjustments, the robot model can be exported from Blender and successfully imported into Unreal Engine. The results can be seen in \autoref{fig:francaScreenshot}

\subsubsection{Mirroring the Robot}
To ensure accurate representation of the physical robot, the digital model must reflect the robot's joint states. A joint state contains the angular position and velocity, and applied effort of a joint. Therefore, a custom class, \texttt{RobotController}, has been developed to facilitate application of the joint state to the rotation axis of the respective link.
Since the rotation occurs around a single axis, both Euler angles and quaternions can be utilized.
Due to a sufficiently high update rate of the joint states, an interpolation between the received frames is not necessary.
By appropriately segmenting and positioning the individual meshes of the model entity,along with correctly defining the rotation axes and angles for each link, an adequate digital counterpart of the physical cobots has been achieved, accurately reflecting their poses.

To closely mirror the physical setup in the virtual environment, both the joystick and the gripper must also be implemented. While the joystick can be considered an extension of the leader's end effector flange, it is also essential for the \ac{dt} to reflect the motion of the gripper's fingers. To achieve this, the gripper joint state, published by the gripper controller from Franka Robotics, is utilized. This topic provides the current distance of the fingers in relation to the fully closed gripper.

The effectiveness of the gripper in the \ac{dt} heavily relies on the tracking method used for grasped objects. To track the assets the \ac{mocap} system is used, which necessitates a minimum object size to accommodate tracking marker installation. These assets are represented by meshes in the \ac{dt} environment.
While infrared tracking systems offer accurate position and velocity data, continuous tracking cannot be guaranteed in all conditions, such as when the tracked object is occluded from multiple cameras.
To address potential tracking interruptions, confirmation from the gripper controller upon a successful grasp provides an additional layer of certainty. This approach ensures that the digital representation of the asset remains accurately aligned with the virtual gripper, even in the event of temporary tracking failures. 
An asset is considered successfully grasped when the gripper's fingers cease movement during the closing action before reaching complete closure. 
 
Upon release of the asset---detected by changes in the gripper's finger position---external position tracking is reactivated. In cases where \ac{mocap} data is unavailable at the moment of release, the physics engine temporarily takes control, simulating the object's motion until reliable tracking data becomes available again. This approach ensures a seamless transition in the \ac{dt} representation, minimizing discrepancies between the virtual and physical environments.

\subsection{Network \& Latency Measurement}
Given that the test environment and \ac{dt} communicate via network, replicating an application, where timing behavior is critical, a performance evaluation of the \ac{dt} is necessary. Consequently, measurements were conducted to assess the latency of the network and the involved processing pipeline.

\begin{figure}[ht]
    \centering
    \includegraphics[width=1.0\linewidth]{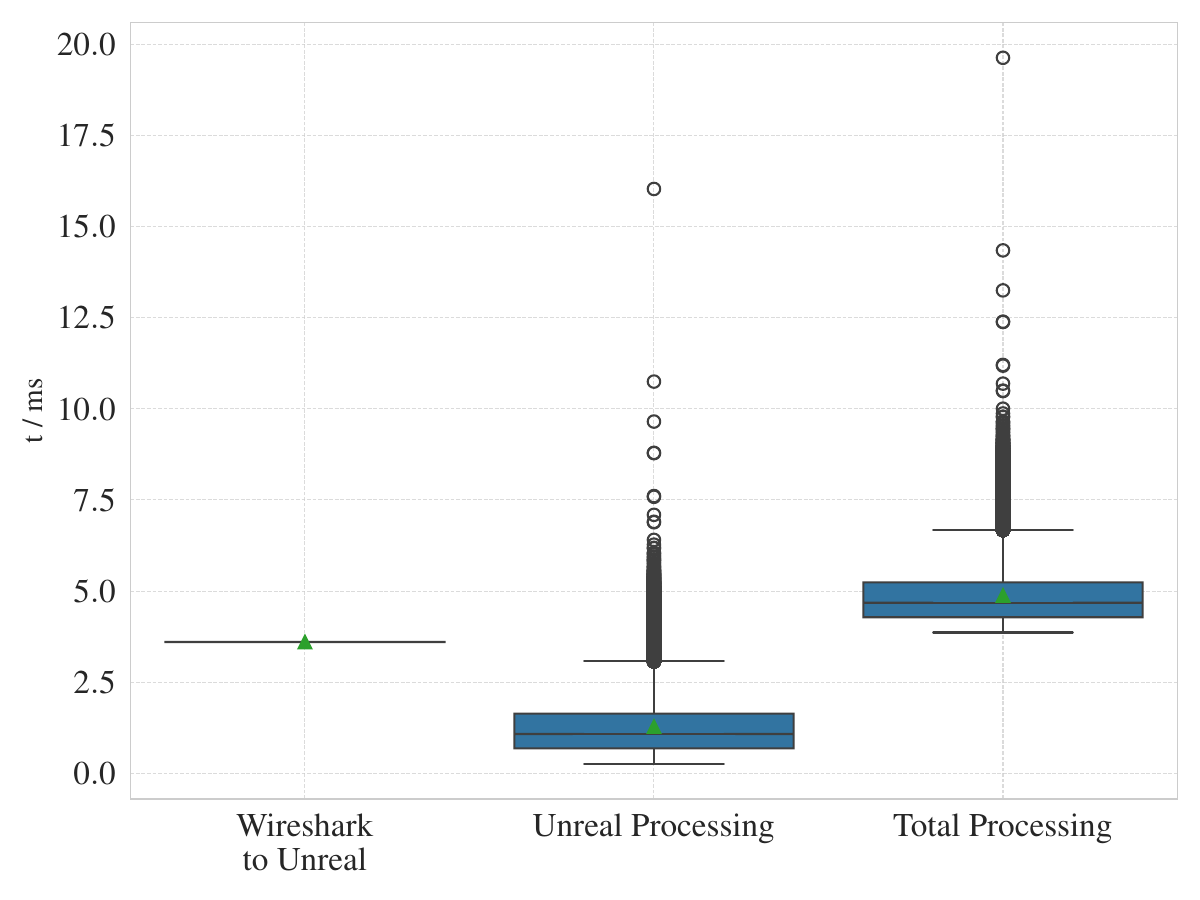}
    \caption{
    Time differences in processing steps from Wireshark to digital twin. n=200k.
    Delta times from Wireshark Capture to receive time in unreal (left). The processing time needed in unreal from receive time and package applied (middle). Combined times (right).}
    \label{fig:deltaBoxplot}
\end{figure}

Latency is evaluated at three stages: i) from the capture of packets in Wireshark their to reception by the UDP socket, ii) from the reception at the UDP socket to the completion of data processing, and iii) the total time encompassing both preceding stages. The results are depicted in \autoref{fig:deltaBoxplot}.

The time delta from packet capture to UDP socket reception demonstrates consistent performance, with a median of \qty{3.6}{\milli\second} and an IQR of \qty{0}{\milli\second}, indicating stable data transmission and socket handling within the system. In contrast, the time delta for data processing exhibits greater variability, with a median of  \qty{1.08}{\milli\second} and an IQR of \qty{0.96}{\milli\second}. Outliers exceeding suggest occasional processing delays.

The combined total processing time reflects these characteristics, with a median of approximately \qty{4.68}{\milli\second}, an IQR of \qty{0.96}{\milli\second}.

Overall performance measurements show frame rates in the range of \SIrange{60}{76}{fps}, corresponding to a frame time of about \SIrange{13}{16}{ms}. In the majority of cases, this frame time is bound by GPU performance rather than CPU or network constraints.

The primary source of delay in the Unreal processing step arises from scheduling latency, which occurs when the UDP socket thread transfers the task to the game thread.

\subsection{Mitigating Tracking Errors}
When marker objects are partially or completely occluded by the operator, a cobot, or nearby objects, the \ac{mocap} system  can loose track of this object, mistakenly identifying it for a different object with similar patterns. 
This results in significant jitter and occasional large, seemingly random shifts in the cobot’s virtual positions. To mitigate jittering positions, a sliding window mean filter is applied to the received coordinate data. Optimal balance between jitter and introducing latency can be provided by a window size in the range from 5 to 15 samples. Additionally, the virtual cobot stations are constrained to rotate solely around the vertical axis (y-axis in Unreal), mirroring the upright orientation of the physical stations. This restriction further minimizes noticeable fluctuations in the virtual positions. For counteracting large and abrupt changes in position, constraints for velocity between two coordinate samples are also implemented.

However, this method has an inherent trade-off. If a marker object moves while it is occluded, the \ac{dt} may interpret its new position as an abrupt shift once tracking resumes. To address this, a check for stable states is introduced by using a sliding buffer storing dropped coordinate frames. Once this buffer has values and new frames consistently fit the new coordinates, this position is accepted as stable.

\subsection{Refinement and Additional Features}

\subsubsection{Prohibited Zones}
The \ac{dt} environment provides the cobots with an understanding of their environment. By incorporating additional real-world objects, and the positions of the cobots, not only the cobots' capabilities can be enhanced, but also safety can be improved.
To enhance safety the concept of prohibited zones is introduced, requiring the cobot to halt its motion upon touching these areas. A vector, oriented outward from the prohibited zone, is computed and transmitted to the respective cobot via a TCP connection. Based on this vector, the robot controller computes a virtual counterforce to prevent the physical cobot from entering the prohibited zone and to ensure that the teleoperation operator can recognize the zone effectively.
This approach enables the \ac{dt} environment to resemble physical world laboratories even more closely and enhance it's capabilities by allowing dynamic bounding constraints, when the respective asset is tracked by the \ac{mocap} system.

\subsubsection{Logging and Replay}
To facilitate retrospective analysis of past events, a Log and Replay functionality is introduced. This feature enables the systematic recording and playback of command sequences, allowing for in-depth evaluation and troubleshooting of system behavior. The data is stored in the Unreal Buffer Archive format, ensuring efficient storage. This approach provides valuable insights for performance optimization, error diagnosis, and iterative system improvements.

\subsubsection{Presentation of the Digital Twin}
A \ac{dt} offers various benefits beyond basic visualization, such as providing enhanced information access and improved safety measures. However, it is essential to consider the optimal platform to represent this virtual model for user interaction.

One approach supported by Unreal Engine is Pixel Streaming. This method initializes a JavaScript server, which allows devices to connect via a designated IP address and port. Once connected, subscribers receive the viewport stream from a selected camera, which can be dynamically adjusted to show different cameras or perspectives. Pixel Streaming supports multiple simultaneous subscribers, enabling various viewing points to be displayed for the \ac{dt} on different devices.

An additional benefit of Pixel Streaming is its built-in support for input devices, allowing to send input events back to the \ac{dt}. These input events are processed by the \ac{dt}, enabling interaction directly from the client device, e.g. to control the view on the \ac{dt}. 
However, Pixel Streaming introduces significant latency, with observed delays exceeding \qty{250}{ms}, even when running on the same system as the \ac{dt} is processed. Therefore, the use of a directly connected display is recommended to minimize latency and ensure a more responsive and seamless visualization.

%% file: chapter/conclusion.tex
\label{sec:conclusion}

This work presents the design and implementation of a \acf{dt} environment integrating robotic arms and an optical infrared tracking system within Unreal Engine 5. The developed framework provides a replicable foundation for advancing \ac{dt} applications in automation and robotics. The proposed solution provides contextual awareness of the environment through infrared tracking while also incorporating a critical safety dimension, such as the designation of prohibited zones for specific assets. The level of \acf{ai} integration into \acp{dt} is flexible and can be individually tailored towards project requirements---whether to utilize deterministic computer vision methods for object tracking or advanced \ac{ai} models for cobot interaction.

This study offers practical guidance on creating robotic entities, enabling video streaming, integration with the \acf{ros2}, and establishing robust bidirectional communication channels. These insights aim to assist researchers and developers in constructing and enhancing \ac{dt} systems for a wide range of applications.

As a direction for future work, exploring \acf{xr} technologies within \ac{dt} environments demonstrates potential for enhancing immersion, enabling precise volumetric interactions, and providing augmented experiences without the need for an additional screen. 
Since \acf{xr} applications demand substantial computational resources, optimizing the scene and meshes to reduce GPU load is a crucial improvement. Additionally, stabilizing scheduling performance by utilizing shared memory instead of passing tasks to the game thread can help minimize latency fluctuations and reduce overhead, particularly in high-load scenarios where the number of entities inherited by the \ac{dt} scales up, leading to an increased volume of incoming packets.

Additionally, a comparative analysis of various graphical engines---including Unity and NVIDIA Omniverse---would help assess their suitability and performance for \ac{dt} implementations in robotics.

In light of recent advancements in \ac{ai}, \acfp{lam} have significantly elevated the importance of \acp{dt}. Implementing a fully functional \ac{lam} capable of dynamic, context-dependent interaction with human operators presents an especially compelling research avenue. In particular novel \ac{ai} supporting tools like NVIDIA Cosmos, can help facilitating such an idea by facilitating the adaption of the robot to complete new settings and environments.

%% file: chapter/acknowledgement.tex
This work has been supported by the Federal Ministry of Education and Research
of the Federal Republic of Germany (Förderkennzeichen 16KISK003K,    
Open6GHub). The authors alone are responsible for the
content of the paper.

%% file: preprint_main.bbl
\begin{thebibliography}{10}
\providecommand{\url}[1]{#1}
\csname url@samestyle\endcsname
\providecommand{\newblock}{\relax}
\providecommand{\bibinfo}[2]{#2}
\providecommand{\BIBentrySTDinterwordspacing}{\spaceskip=0pt\relax}
\providecommand{\BIBentryALTinterwordstretchfactor}{4}
\providecommand{\BIBentryALTinterwordspacing}{\spaceskip=\fontdimen2\font plus
\BIBentryALTinterwordstretchfactor\fontdimen3\font minus
  \fontdimen4\font\relax}
\providecommand{\BIBforeignlanguage}[2]{{%
\expandafter\ifx\csname l@#1\endcsname\relax
\typeout{** WARNING: IEEEtran.bst: No hyphenation pattern has been}%
\typeout{** loaded for the language `#1'. Using the pattern for}%
\typeout{** the default language instead.}%
\else
\language=\csname l@#1\endcsname
\fi
#2}}
\providecommand{\BIBdecl}{\relax}
\BIBdecl

\bibitem{OmniverseShowCase}
\BIBentryALTinterwordspacing
Nvidia. (2024) Omniverse digital twin showcase. [Online]. Available:
  \url{https://www.nvidia.com/de-de/omniverse/solutions/digital-twins/}
\BIBentrySTDinterwordspacing

\bibitem{UnityShowCase}
\BIBentryALTinterwordspacing
Unity. (2024) Unity digital twin showcase. [Online]. Available:
  \url{https://unity.com/de/topics/digital-twin-definition}
\BIBentrySTDinterwordspacing

\bibitem{UnrealShowCase}
\BIBentryALTinterwordspacing
EPIC. (2024) Unreal digital twin showcase. [Online]. Available:
  \url{https://www.unrealengine.com/en-US/digital-twins}
\BIBentrySTDinterwordspacing

\bibitem{MASHALY2021299}
M.~Mashaly, ``Connecting the twins: A review on digital twin technology \& its
  networking requirements,'' \emph{Procedia Computer Science}, vol. 184, pp.
  299--305, 2021.

\bibitem{HALUSKOVA20231471}
B.~Halúsková, ``Digital twin in smart city,'' \emph{Transportation Research
  Procedia}, vol.~74, pp. 1471--1478, 2023, tRANSCOM 2023: 15th International
  Scientific Conference on Sustainable, Modern and Safe Transport.

\bibitem{Petershans2024_RoboticTeleoperationRealWorld}
J.~Petershans, J.~Herbst, M.~Rüb, E.~Mittag, and H.~D. Schotten, ``Robotic
  {{Teleoperation}}: {{A Real-World Test Environment}} for {{6G
  Communications}},'' in \emph{28. {{ITG-Fachtagung
  Mobilkommunikation}}}.\hskip 1em plus 0.5em minus 0.4em\relax VDE-Verlag, May
  2024, pp. 106--111.

\bibitem{Coronado2023}
E.~Coronado, S.~Itadera, and I.~G. Ramirez-Alpizar, ``Integrating virtual,
  mixed, and augmented reality to human robot interaction applications using
  game engines: A brief review of accessible software tools and frameworks,''
  \emph{Applied Sciences}, vol.~13, no.~3, 2023.

\bibitem{10188847}
H.~Xu, J.~Wu, Q.~Pan, X.~Guan, and M.~Guizani, ``A survey on digital twin for
  industrial internet of things: Applications, technologies and tools,''
  \emph{IEEE Communications Surveys \& Tutorials}, vol.~25, no.~4, pp.
  2569--2598, 2023.

\bibitem{ITU-T}
\BIBentryALTinterwordspacing
Itu telecommunication standardization sector (e.g. dt networks). [Online].
  Available: \url{https://www.itu.int/en/ITU-T/}
\BIBentrySTDinterwordspacing

\bibitem{10633478}
R.~Gu, T.~Barbuceanu, N.~Xiong, and T.~Seceleanu, ``Experiences in building a
  digital twin framework: Challenges and possible solutions,'' in \emph{2024
  IEEE 48th Annual Computers, Software, and Applications Conference (COMPSAC)},
  2024, pp. 531--536.

\bibitem{10717667}
C.~G. Zarco, ``Digital twin and services: A theoretical development
  framework,'' in \emph{2024 IEEE Technology and Engineering Management Society
  (TEMSCON LATAM)}, 2024, pp. 1--6.

\bibitem{10241448}
X.~Yang, L.~Yang, and S.~Li, ``A digital twin application framework in the
  field of industry,'' in \emph{2023 IEEE 18th Conference on Industrial
  Electronics and Applications (ICIEA)}, 2023, pp. 1615--1619.

\bibitem{krummacker_digital_2023}
D.~Krummacker, M.~Reichardt, C.~Fischer, and H.~D. Schotten, ``Digital twin
  development: Mathematical modeling,'' in \emph{2023 IEEE 6th International
  Conference on Industrial Cyber-Physical Systems (ICPS)}, 03 2023, pp. 1--8.

\bibitem{kritzinger_digital_2018-3}
W.~Kritzinger, M.~Karner, G.~Traar, J.~Henjes, and W.~Sihn, ``Digital {Twin} in
  manufacturing: {A} categorical literature review and classification,''
  \emph{IFAC-PapersOnLine}, vol.~51, no.~11, pp. 1016--1022, 2018.

\bibitem{singh_digital_2021}
M.~Singh, E.~Fuenmayor, E.~P. Hinchy, Y.~Qiao, N.~Murray, and D.~Devine,
  ``\BIBforeignlanguage{en}{Digital {Twin}: {Origin} to {Future}},''
  \emph{\BIBforeignlanguage{en}{Applied System Innovation}}, vol.~4, no.~2,
  p.~36, Jun. 2021.

\bibitem{cruz_systematic_2022}
R.~J. M.~d. Cruz and L.~A. Tonin, ``Systematic review of the literature on
  {Digital} {Twin}: a discussion of contributions and a framework proposal,''
  \emph{Gestão \& Produção}, vol.~29, p. 9621, 2022.

\bibitem{negri_clarifying_2023}
E.~Negri and T.~A. Abdel-Aty, ``Clarifying concepts of {Metaverse}, {Digital}
  {Twin}, {Digital} {Thread} and {AAS} for {CPS}-based production systems,''
  \emph{IFAC-PapersOnLine}, vol.~56, no.~2, pp. 6351--6357, Jan. 2023.

\bibitem{zhang_literature_2024}
Q.~Zhang, J.~Liu, and X.~Chen, ``\BIBforeignlanguage{en}{A {Literature}
  {Review} of the {Digital} {Thread}: {Definition}, {Key} {Technologies}, and
  {Applications}},'' \emph{\BIBforeignlanguage{en}{Systems}}, vol.~12, no.~3,
  p.~70, Mar. 2024.

\bibitem{Kapusuzoglu2020_PhysicsInformedHybridMachine}
B.~Kapusuzoglu and S.~Mahadevan, ``Physics-{{Informed}} and {{Hybrid Machine
  Learning}} in {{Additive Manufacturing}}: {{Application}} to {{Fused Filament
  Fabrication}},'' \emph{JOM}, vol.~72, no.~12, pp. 4695--4705, Dec. 2020.

\bibitem{Daw2022_PhysicsGuidedNeuralNetworks}
A.~Daw, A.~Karpatne, W.~D. Watkins, J.~S. Read, and V.~Kumar,
  \emph{Physics-{{Guided Neural Networks}} ({{PGNN}}): {{An Application}} in
  {{Lake Temperature Modeling}}}, 1st~ed.\hskip 1em plus 0.5em minus
  0.4em\relax Boca Raton: {Chapman and Hall/CRC}, Jun. 2022, pp. 353--372.

\bibitem{__SpecificationAssetAdministration}
\BIBentryALTinterwordspacing
Specification of the asset administration shell part 1: Metamodel – {IDTA}
  number: 01001-3-0-1. [Online]. Available:
  \url{https://industrialdigitaltwin.org/content-hub/aasspecifications/part1_metamodel}
\BIBentrySTDinterwordspacing

\bibitem{mike_rauh.etal_2022_AIAssetManagement}
L.~Rauh, M.~Reichardt, and H.~D. Schotten, ``{AI} asset management: a case
  study with the asset administration shell ({AAS}),'' in \emph{2022 {IEEE}
  27th International Conference on Emerging Technologies and Factory Automation
  ({ETFA})}.\hskip 1em plus 0.5em minus 0.4em\relax {IEEE}, pp. 1--8.

\bibitem{Coronado2024_PathIndustry50}
E.~Coronado, T.~Ueshiba, and I.~G. {Ramirez-Alpizar}, ``A path to {{Industry}}
  5.0 {{Digital Twins}} for {{Human}}--{{Robot Collaboration}} by {{Bridging
  NEP}}+ and {{ROS}},'' \emph{Robotics}, vol.~13, no.~2, p.~28, Feb. 2024.

\bibitem{phobos}
K.~von Szadkowski and S.~Reichel, ``Phobos: A tool for creating complex robot
  models,'' \emph{Journal of Open Source Software}, vol.~5, no.~45, p. 1326,
  2020.

\end{thebibliography}
